\documentclass[11pt,a4paper]{article}
\usepackage[hyperref]{emnlp2020}
\usepackage{times}
\usepackage{latexsym}
\usepackage{amssymb, amsmath, amsxtra, bbold, isomath, mathdots, stmaryrd, wasysym}
\usepackage{todonotes, array, csquotes}

\usepackage{microtype}

\aclfinalcopy 
\def\aclpaperid{2285} 

\title{Knowing What You Know: Calibrating Dialogue Belief State Distributions via Ensembles}

\author{Carel van Niekerk, Michael Heck, Christian Geishauser \\
  \textbf{Hsien-Chin Lin, Nurul Lubis, Marco Moresi, Milica Gašić} \\
  Heinrich Heine University Düsseldorf, Germany \\
  \texttt{niekerk,heckmi,geishaus,linh,lubis,moresi,gasic@hhu.de}}

\date{}

\begin{document}
\maketitle
\begin{abstract}
The ability to accurately track what happens during a conversation is essential for the performance of a dialogue system. Current state-of-the-art multi-domain dialogue state trackers achieve just over $55\%$ accuracy on the current go-to benchmark, which means that in almost every second dialogue turn
they place full confidence in an incorrect dialogue state. 
Belief trackers, on the other hand, maintain a distribution over possible dialogue states. However, they lack in performance compared to dialogue state trackers, and do not produce well calibrated distributions. In this work we present state-of-the-art performance in calibration for multi-domain dialogue belief trackers using a calibrated ensemble of models. Our resulting dialogue belief tracker also outperforms previous dialogue belief tracking models in terms of accuracy.
\end{abstract}

\section{Introduction}
Task-oriented dialogue systems aim to act as assistants to their users, solving tasks such as finding a restaurant, booking a train, or providing information about a tourist attraction. They have become very popular with the introduction of virtual assistants such as Siri and Alexa.

Two tasks are fundamental to such a system. The first is the ability to track what happened in the conversation, referred to as \textbf{tracking}. Based on the result of tracking, the system needs to conduct the conversation towards the fulfilment of the user goal, referred to as \textbf{planning}. The tracking component summarises the dialogue history, or the past, while the planning component manages the dialogue and concerns the future. In this work we focus on the first component.

Early approaches to statistical dialogue modelling view dialogue as a Markov decision process~\citep{levin1998mdp} and define a set of dialogue states that the conversation can be in at any given dialogue turn. The tracking component tracks the \textbf{dialogue state}. 
In recent years
discriminative models achieve state-of-the-art dialogue state tracking (DST) results~\citep{kim2019somdst,zhang2019dsdst,heck2020trippy}. Still, in a multi-domain setting such as MultiWOZ~\citep{eric2019multiwoz,budzianowski2018large}, they achieve an accuracy of just over $55\%$. This means that in approximately $45\%$ of cases they make a wrong prediction and, even worse, they have full confidence in that wrong prediction.

In the wake of statistical dialogue modeling, the use of partially observable Markov decision processes has been proposed to address this issue. The idea is to model the probability over all possible dialogue states in every dialogue turn~\cite{williams2007pomdp}. This probability distribution is referred to as the \textbf{belief state}. The advantages of belief tracking are probably best illustrated by an excerpt from a dialogue with a real user in \cite{metallinou2013dst}: even though the dialogue state predicted with the highest probability is not the true one, the system is able to provide a valid response because the true dialogue state also has assigned a non-zero probability.

\begin{displayquote}
A model is considered well \textbf{calibrated} if its confidence estimates are aligned with the empirical likelihood of its predictions \citep{desai2020calibration}.
\end{displayquote}

The belief state can be modelled by deep learning-based approaches such as the neural belief tracker~\citep{mrksic-etal-2017-neural}, the multi-domain belief tracker~\citep{ramadan2018mdbt}, the globally conditioned encoder belief tracker~\citep{nouri2018gce} and the slot utterance matching belief tracker~(SUMBT)~\citep{lee2019sumbt} models. None of these models however address the issue of calibrating the probability distribution that they provide, resulting in them being more confident than they should be. In a dialogue setting, overconfidence can lead to bad decisions and unsuccessful dialogues.

In this work, we present methods for learning well-calibrated belief distributions. Our contributions are the following:
\begin{itemize}
\item We present the state-of-the-art performance in calibration for dialogue belief trackers using a calibrated ensemble of models, called the calibrated ensemble belief state tracker (CE-BST).
\item Our model achieves best overall joint goal accuracy among the state-of-the-art \textbf{belief} tracking models.
\end{itemize}
Such a well-calibrated belief tracking model is essential for the planning component to successfully conduct dialogue. 

\section{Related Work}
Since no other belief tracking methods that we are aware of have achieved success in producing well-calibrated confidence, we look towards methods used in other language tasks. Natural language inference is a related task that also benefits from well-calibrated confidence in predictions. \citet{desai2020calibration} introduce the use of post-processing techniques such as temperature scaling to produce better-calibrated confidence estimates.

Additionally, there have been recent advances in the construction of more adequate loss functions. These methods, including Bayesian matching and prior networks, aim to learn well-calibrated models without the burden of requiring many extra parameters. These methods achieve good calibration in computer vision tasks such as CIFAR \citep{joo2020bm,malinin2018prior,szegedy2016ls}.

When the limitations of a single model still inhibit us from producing more accurate and better-calibrated models, a popular alternative is to use an ensemble of models. Recently \citet{malinin2020uncertainty} showed the success of using an ensemble of models for machine translation, and in particular utilising accurate confidence predictions for analysing translation quality.

\section{Calibration Techniques}
In this section we explain the details of three calibration techniques that we apply to dialogue belief tracking.
\subsection{Loss Functions}
The loss function can have a great impact on the calibration and accuracy of models. The most commonly used loss function in belief tracking is the standard softmax cross entropy loss. However, it tends to cause overconfident predictions where most of the probability is placed on the top class.

Label smoothing cross entropy \citep{szegedy2016ls} aims to resolve this problem by replacing the one-hot targets of cross entropy with a smoothed target distribution. That is, for label $y_i$ and smoothing parameter $\alpha \in \left( 0,\frac{1}{K} \right]$, the target distribution will be:
\begin{equation}
t(c|\alpha, y_i)
= \begin{cases} 
      1 - (K-1)\alpha & c=y_i, \\
      \alpha & \mathrm{otherwise},
   \end{cases}
\end{equation}
where $K$ is the number of possible values of $c$. The loss for a model with parameters $\boldsymbol{\theta}$ and a set of $N$ output logits $\hat{\mathbf{z}}_1, \hat{\mathbf{z}}_2, ..., \hat{\mathbf{z}}_N$ with true labels $y_1, y_2, ..., y_N$ is defined as:
\begin{equation}
\mathcal{L}(\boldsymbol{\theta}, \alpha) = \frac{1}{N} \sum_{i=1}^{N} \mathbf{KL} \left[ \mathrm{Softmax}(\hat{\mathbf{z}}_i) || t(c_i|\alpha, y_i) \right],
\end{equation}
where $\mathbf{KL}$ is the Kullback–Leibler divergence between two distributions~\citep{kullback1951information}.

Alternatively, Bayesian matching loss \citep{joo2020bm} uses a Dirichlet distribution as the final activation function. The target is constructed using the Bayes rule, where we assume the observed label $y_i$ to be an observation from a categorical distribution $y_i|\boldsymbol{\pi}_i \sim \mathrm{Cat}(\boldsymbol{\pi}_i)$ and $\boldsymbol{\pi}_i$ is the true underlying distribution of the label. To introduce uncertainty into the target distribution we assume that the prior of $\boldsymbol{\pi}_i$ is a Dirichlet distribution, $\mathrm{Dir}(\mathbf{1})$. In this way, we have a highly uncertain prior distribution. From this it can be shown that the posterior will be $\boldsymbol{\pi}_i | y_i \sim \mathrm{Dir}(\mathbf{1} + \mathbf{I}(y_i))$, where $\mathbf{I}(y_i)$ is the one-hot representation of $y_i$.
The loss function is then constructed using the negative log likelihood of the true label given the predicted distribution $\hat{\boldsymbol{\pi}}_i \sim \mathrm{Dir}(\hat{\mathbf{z}}_i)$, penalised by the KL divergence from the the uncertain $\mathrm{Dir}(\mathbf{1})$ distribution:
\begin{multline}
\mathcal{L}(\boldsymbol{\theta}, \lambda) = \sum_{i=1}^{N} \{ \lambda \mathbf{KL} \left[ \hat{\boldsymbol{\pi}}_i || \mathrm{Dir}(\mathbf{1})\right] - \\
\mathbb{E}_{\hat{\boldsymbol{\pi}}_i} \left[ \mathrm{log}( p(y_i|\hat{\boldsymbol{\pi}}_i))\right] \},
\end{multline}
where $\lambda>0$ is the penalisation parameter.

\subsection{Ensemble Distribution Estimation}
\label{section:ensemble}
From a Bayesian viewpoint, the probability of observing an outcome given the observed examples can be broken down into two components: the predictive distribution of the model and the posterior of the model given the observed examples. The posterior of the model given the data is an unknown distribution which can be estimated in various ways. One method is to use an ensemble of models, where the ensemble acts as an estimator for the posterior distribution of the parameters, $p(\boldsymbol{\theta} | \mathcal{D})$, where $\mathcal{D}$ represents the observed examples. Let $q(\boldsymbol{\theta})$ represent the distribution over all possible members of an ensemble. This distribution could be seen as the ensemble estimate of the posterior, $p(\boldsymbol{\theta} | \mathcal{D})$, \citep{malinin2019ensemble,malinin2020uncertainty}. Hence,
\begin{equation}
\hat{p}( y | \mathbf{x}, \mathcal{D} ) = \int p( y | \mathbf{x}, \boldsymbol{\theta} ) q( \boldsymbol{\theta} ) \mathrm{d}\boldsymbol{\theta}.
\end{equation}
Since this integral is still intractable we need to estimate it using Monte Carlo. To sample from the ensemble distribution $q( \boldsymbol{\theta} )$ we consider two approaches: using dropout during inference to collect an ensemble of $N$ equally likely models \citep{gal2016dropout}, or alternatively bootstrap sampling $N$ equally likely subsets of the data to train $N$ equally likely ensemble members. Let these $N$ members be $\{ \boldsymbol{\theta}^{(1)}, \boldsymbol{\theta}^{(2)}, ..., \boldsymbol{\theta}^{(N)} \}$. The estimated predictive distribution can then be calculated as follows:
\begin{equation}
\hat{p}( y | \mathbf{x}, \mathcal{D} ) = \frac{1}{N} \sum_{i=1}^{N} p( y | \mathbf{x}, \boldsymbol{\theta}^{(i)})
\end{equation}

\subsection{Temperature Scaling}
Temperature scaling is a post-processing technique which scales the logits of the model by a scaling factor $\beta > 1$ \citep{guo2017temp}, resulting in better-calibrated estimates. The temperature scaling parameter $\beta$ can be trained on a development set.

\section{Experimental Setup}
\label{ssec:nbt_results}
We seek to build a well-calibrated dialogue belief tracker. For our baseline belief tracker, we use the SUMBT model architecture \cite{lee2019sumbt}, which uses BERT \citep{devlin2018bert} as a turn encoder and multi-head attention for slot candidate matching. We perform all experiments on the MultiWOZ 2.1 dataset~\citep{eric2019multiwoz}, the current standard dataset for multi-domain dialogue. When training using Bayesian matching, we use a scaling coefficient of $\lambda=0.003$, and for label smoothing, a smoothing coefficient of $\alpha=0.05$. For the ensemble belief tracker, we train $10$ identical independent models, each with a sub-sample of $7500$ dialogues. All hyper-parameters are obtained using a parameter search based on validation set performance.
For all training, we use the BERT-base-uncased model from PyTorch Transformers \citep{Wolf2019HuggingFacesTS} for turn embedding. We use a gated recurrent unit with a hidden dimension $300$ for latent tracking and Euclidean distance for value candidate scoring.
During training, we use a learning rate of $5e-5$ in combination with a linear learning rate scheduler, the warm-up proportion is set to $0.1$. A dropout rate of $0.3$ is used, and training is performed for $100$ epochs.\footnote{Our code will be made available at \url{https://gitlab.cs.uni-duesseldorf.de/general/dsml/calibrating-dialogue-belief-state-distributions}.}

\section{Evaluation Metrics}
\subsection{Joint Goal Accuracy}
The joint goal accuracy (JGA) is the percentage of turns for which the model predicts the complete user goal correctly. We further propose the introduction of an adjusted top $3$ JGA, which considers a user goal prediction correct if the true label for each slot is among the top $3$ predicted candidates for that slot in the belief state given there are at least $5$ possible candidates.
\subsection{L2 Norm Error}
The L2 norm error is the L2 norm of the difference between the true labels and the predicted distributions. To form the user goals and belief states we concatenate all the slot labels and slot distributions. This error measure does not only consider the accuracy of the predictions but also the uncertainty.
\subsection{Joint Goal Calibration Error}
A well-calibrated model is one where the accuracy is aligned with the confidence predictions. The expected calibration error (ECE) evaluates the calibration by measuring the difference between the model's confidence and accuracy \citep{guo2017temp}, meaning a lower ECE indicates better calibration. Hence:
\begin{equation}
\mathrm{ECE} = \sum_{k=1}^B \frac{b_k}{N} |\mathrm{acc}(k) - \mathrm{conf}(k)|,
\end{equation}
where $B$ is the number of bins, $b_k$ are the bin sizes, $N$ the number of observations, $\mathrm{acc}(k)$ and $\mathrm{conf}(k)$ the accuracy and confidence measures of bin $k$.
We also propose an adapted ECE, called the expected joint goal calibration error (EJCE), which uses the joint goal accuracy for bin $k$ as $\mathrm{acc}(k)$, and the following metric as confidence:
\begin{equation}
\mathrm{conf(k)} = \frac{1}{b_k} \sum_{i=1}^{b_k} \min_{s \in \mathrm{slots}} \max_{v \in \mathrm{values}} \hat{p}_i(v|s),
\end{equation}
where $\hat{p}_i(v|s)$ is the predicted probability of value $v$ for slot $s$ given the $i^{th}$ observation in bin $k$.

\section{Results}
\begin{table}[h!]
    \centering
    \begin{tabular}{|m{3.2cm}|m{0.9cm}|m{0.9cm}|m{0.8cm}|}
        \hline
        \small{Model} & \small{JGA} & \small{Top~$3$ JGA} & \small{EJCE} \\
        \hline
        \small{Cross entropy} & \small{$46.78\%$} & \small{$69.97\%$} & \small{$1.996$} \\
        \small{Label smoothing} & \small{$46.32\%$} & \small{$74.57\%$} & \small{$1.292$} \\
        \small{Bayesian matching} & \small{$31.03\%$} & \small{$45.16\%$} & \small{$4.922$} \\
        \hline
        \multicolumn{4}{|c|}{\small{Temperature scaling}} \\
        \hline
        \small{Cross entropy ($1.73$*)} & \small{$46.78\%$} & \small{$69.97\%$} & \small{$4.758$} \\
        \small{Label smoothing ($1.00$*)} & \small{$46.32\%$} & \small{$74.57\%$} & \small{$1.292$} \\
        \hline
        \multicolumn{4}{|c|}{\small{Dropout ensembles}} \\
        \hline
        \small{Cross entropy ($35$**)} & \small{$47.18\%$} & \small{$71.14\%$} & \small{$2.909$} \\
        \small{Label smoothing ($35$**)} & \small{$46.36\%$} & \small{$76.12\%$} & \small{$2.217$} \\
        \hline
        \multicolumn{4}{|c|}{\small{Bootstrap model ensembles}} \\
        \hline
        \small{Label smoothing ($10$**)} & \small{$\mathbf{48.41\%}$} & \small{$\mathbf{84.08\%}$} & \small{$\mathbf{0.841}$} \\
        \hline
    \end{tabular}
    \centering
    \caption{Calibration strategy performance. *temperature scaling coefficient **ensemble size.}
    \label{tab:jga_comparison}
\end{table}
\begin{table}[h!]
    \centering
    \begin{tabular}{|c|c|c|}
        \hline
        \small{Model} & \small{JGA} & \small{L2 Norm} \\
        \hline
        \small{SUMBT~\citep{lee2019sumbt}}& \small{$46.78\%$} & \small{$1.1075$} \\
        \small{CE-BST~(ours)} & \small{$48.41\%$} & \small{$\mathbf{1.1041}$} \\
        \small{SOTA DST} & \small{$<\mathbf{56.0\%}$} & \small{$>1.2445$} \\
        \hline
    \end{tabular}
    \caption{MultiWOZ $2.1$ performance.}
    \label{tab:mwoz}
\end{table}
\begin{figure}[h!]
    \centering
    \includegraphics[scale=0.2]{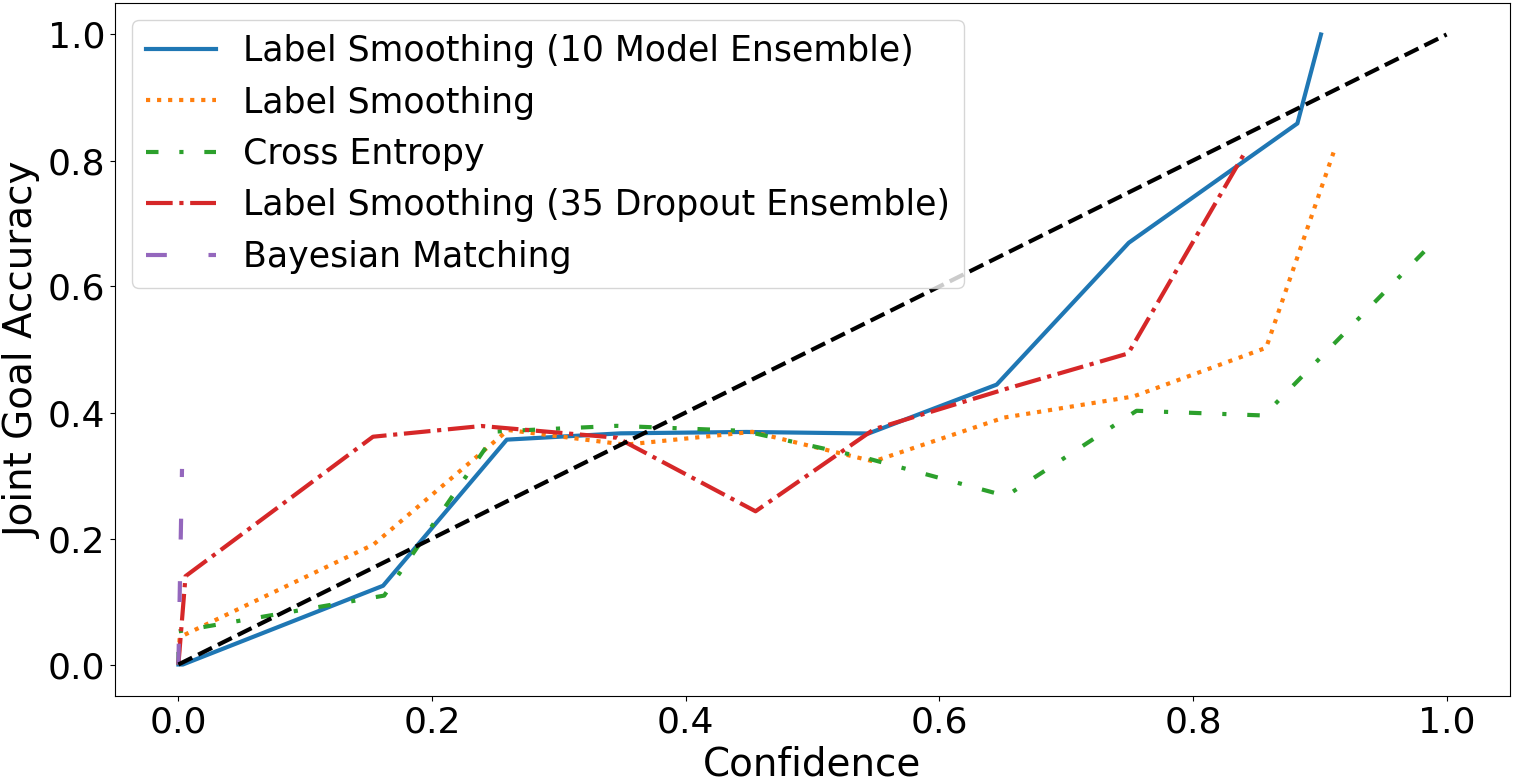}
    \caption{Reliability Diagram.\label{fig:jga_comparison}}
\end{figure}
All of the calibration techniques presented above can be combined. Here, we focus on the most important combinations and present the results in Table~\ref{tab:jga_comparison}. We make the following observations. First, cross entropy on its own leads to a high EJCE, as expected. Second, label smoothing reduces EJCE while leading to a negligible drop in accuracy. Third, Bayesian matching underperformed in our experiments, suggesting a difficulty in choosing the right priors. Fourth, temperature scaling is not an effective way of calibrating uncertainty, as the same calibration is applied to each observation. Finally, the ensemble methods produce very promising results for both accuracy and calibration of the model. In particular, if we look at the Top~$3$ JGA, our method achieves an improvement of $14.11$ percentage points over the baseline, in the Appendix we include a comprehensive set of Top $n$ JGA results. In Figure~\ref{fig:jga_comparison} we plot JGA as a function of confidence. The best calibrated model is the one that is closest to the diagonal, i.e.\ the one whose confidence for each dialogue state is closest to the achieved accuracy. From this reliability diagram we see that both the dropout and model ensembles improve model calibration and do not produce over-confident output as the cross entropy baseline does.
In Table~\ref{tab:mwoz} we compare our model to some of the best performing belief and state tracking models. Here we see that we outperform the best performing \textbf{belief} tracker but the state-of-the-art (SOTA) \textbf{state} trackers~\cite{heck2020trippy, chen2020sst, hosseini2020simpletod} have a significantly higher JGA. However, when analysing the L2 norm\footnote{For a model with a given JGA we can calculate the minimum L2 that such a model can possibly achieve by assuming that it never predicts more than one slot incorrectly.} we see that the uncertainty estimates of \textbf{belief} tracking models compensate for the lower joint goal accuracy. This corroborates our premise that it is important to have well calibrated confidence estimates and not just a high JGA.

\section{Conclusion}

We applied a number of calibration techniques to a baseline dialogue belief tracker. We showed that a label smoothed trained ensemble provides state-of-the-art calibration of the belief state distributions and has the best accuracy among the available \textbf{belief} trackers. Although it does not compete with \textbf{state} trackers in terms of JGA, when considering top 3 predictions it achieves 84.08\% accuracy (Top~3 JGA), almost 30 percentage points above state-of-the art state trackers. 
We also find that our model has the best L2 norm performance, which suggests that the quality of predicted uncertainty is as important as the average JGA. 

It is important to note that the proposed calibration methods can be applied to any neural dialogue belief tracking method. The uncertainty estimates predicted by this model could improve the success of dialogue systems because this model can provide the dialogue manager with a good measure of confidence. This could allow the system to ask questions in moments of confusion. In the Appendix we include example dialogues to illustrate this. In future, we aim to combine the state-of-the-art dialogue state tracking and belief tracking methods to create a method that can achieve both states-of-the-art joint goal accuracy and well-calibrated belief states.

\section*{Acknowledgements}
C. van Niekerk, M. Heck and N. Lubis are supported by funding provided by the Alexander von Humboldt Foundation in the framework of the Sofja Kovalevskaja Award endowed by the Federal Ministry of Education and Research, while C. Geishauser, H-C. Lin and M. Moresi are supported by funds from the European Research Council (ERC) provided under the Horizon 2020 research and innovation programme (Grant agreement No. STG2018 804636).

\bibliographystyle{acl_natbib}
\bibliography{refs}

\appendix

\section{Appendices}
\label{sec:appendix}
\subsection{Joint Goal Accuracy Analysis}
In Table \ref{tab:topn} we compare SUMBT and our CE-BST method using $5$ different top $n$ joint goal accuracy's.
\begin{table}[h!]
    \centering
    \begin{tabular}{|p{1.1cm}|p{0.85cm}|p{0.85cm}|p{0.85cm}|p{0.85cm}|p{0.85cm}|}
        \hline
        \small{Model} & \small{Top $1$} & \small{Top $2$} & \small{Top $3$} & \small{Top $4$} & \small{Top $5$} \\
        \hline
        \small{SUMBT} & \small{$46.78\%$} & \small{$64.61\%$} & \small{$69.97\%$} & \small{$72.10\%$} & \small{$73.70\%$} \\
        \small{CE-BST} & \small{$48.41\%$} & \small{$77.25\%$} & \small{$84.08\%$} & \small{$85.84\%$} & \small{$86.93\%$} \\
        \hline
    \end{tabular}
    \caption{Top $n$ joint goal accuracy comparison.}
    \label{tab:topn}
\end{table}

\subsection{Example Dialogues}
\begin{figure}[h!]
    \centering
    \includegraphics[scale=0.45]{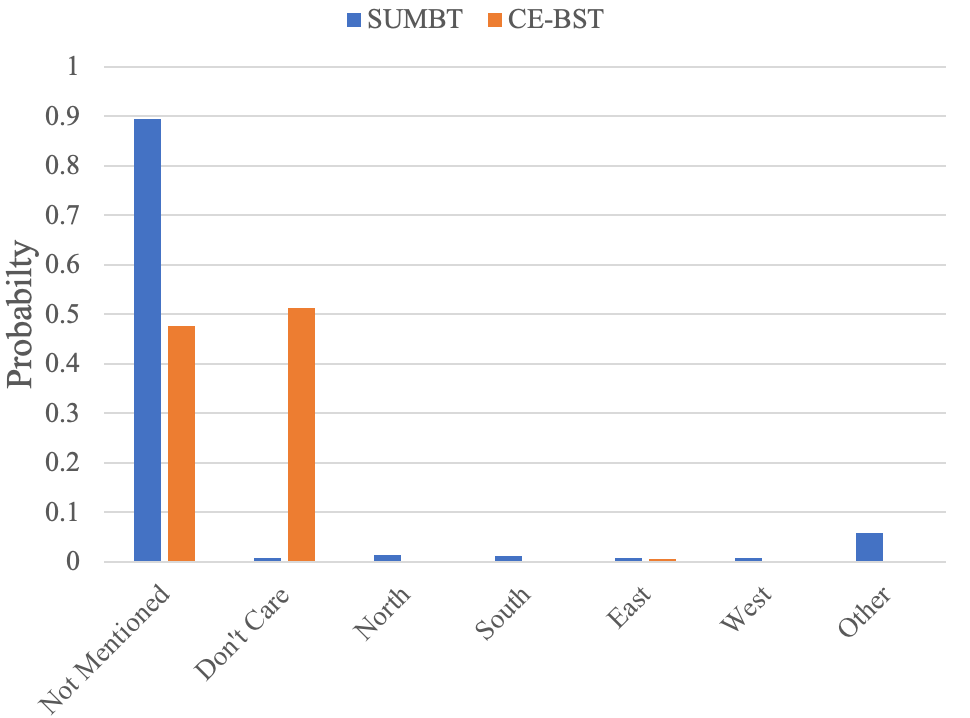}
    \caption{\textit{PMUL3364} Hotel - Location belief state distribution.}
    \label{fig:bs3}
\end{figure}

\begin{figure}[h!]
    \centering
    \includegraphics[scale=0.45]{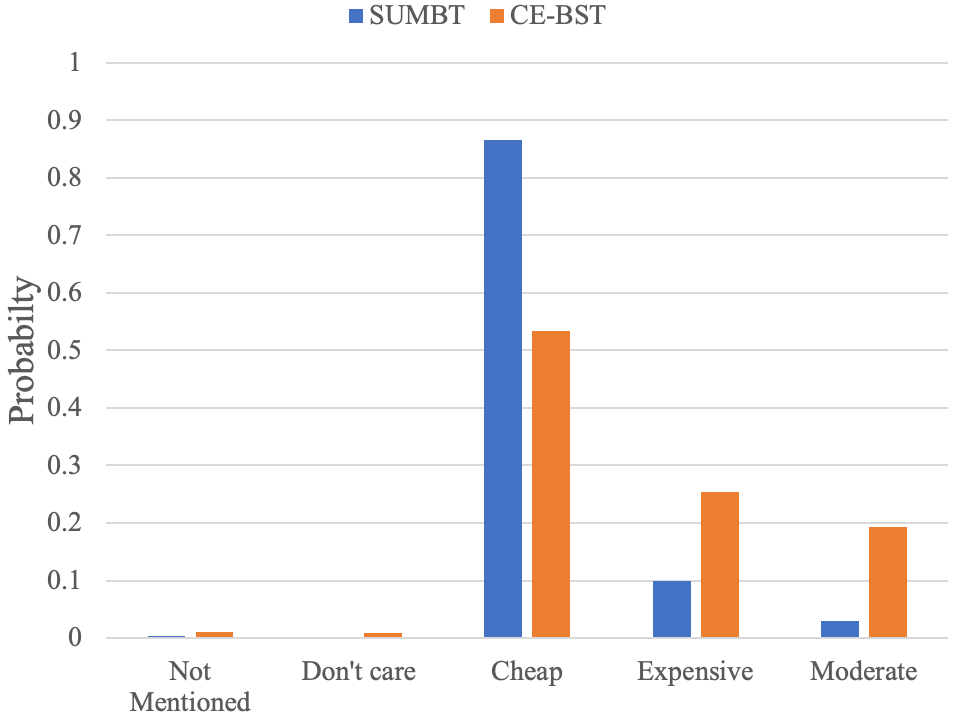}
    \caption{\textit{PMUL4258} Restaurant - Price Range belief state distribution.}
    \label{fig:bs0}
\end{figure}

\begin{figure}[h!]
    \centering
    \includegraphics[scale=0.45]{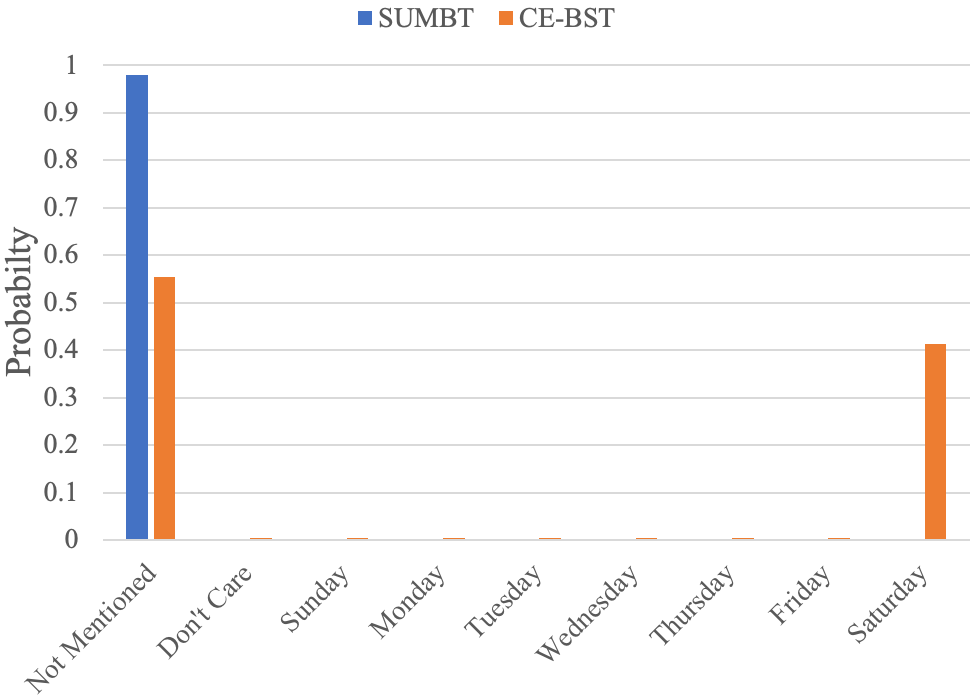}
    \caption{\textit{PMUL4605} Hotel - Book Day belief state distribution.}
    \label{fig:bs}
\end{figure}

\begin{figure}[h!]
    \centering
    \includegraphics[scale=0.45]{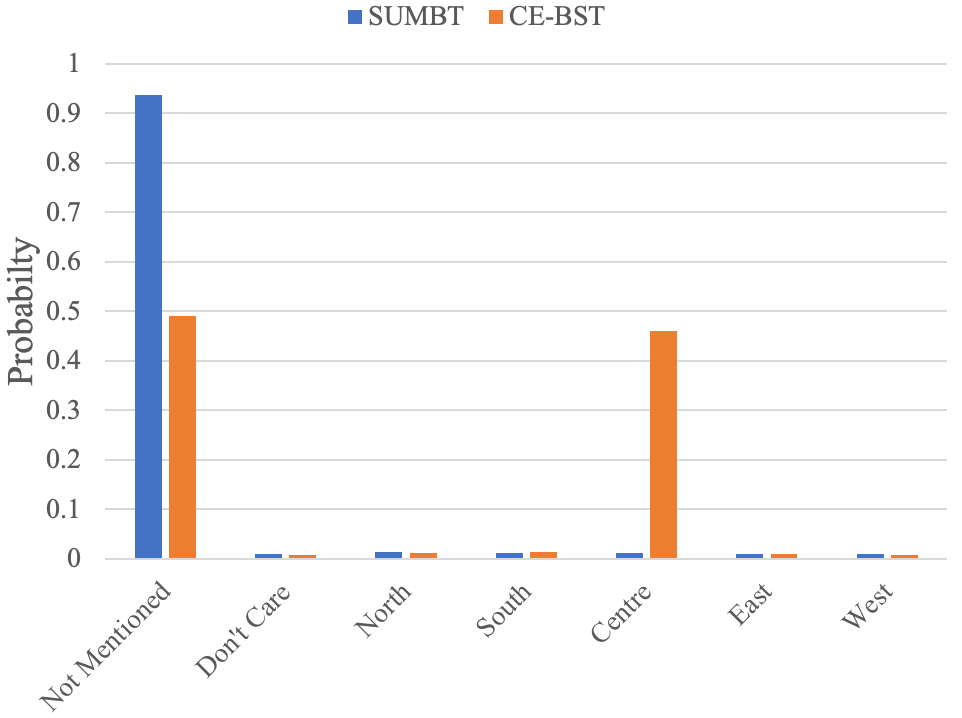}
    \caption{\textit{PMUL3625} Restaurant - Location belief state distribution.}
    \label{fig:bs2}
\end{figure}

In Figures \ref{fig:dial_eg3} - \ref{fig:bs2} we present some example dialogues together with an extract from their belief state distributions. These examples show situations where a well-calibrated belief state distribution could be beneficial for decision making.
\begin{figure}[h!]
    \centering
    \fbox{\begin{minipage}{19em}
    \textbf{User:} I need a place to stay.
    
    \textbf{System:} Sure. I'll need a little more information. Is there an area you are interested in?
    
    \textbf{User:} No specific area. I would like it to be in the moderate price range and it should have free parking. I would also like it to have 4 stars.
    \end{minipage}}
    \caption{Dialogue \textit{PMUL3364} from the MultiWOZ $2.1$ corpus.}
    \label{fig:dial_eg3}
\end{figure}
\begin{figure}[h!]
    \centering
    \fbox{\begin{minipage}{20em}
    \textbf{User:} Can you help me find a place to go in the centre?
    
    \textbf{System:} I can help you with that. Is there a certain kind of attraction that you would like to visit?
    
    \textbf{User:} Surprise me! Give me the postcode as well.
    
    \textbf{System:} Would you prefer the castle galleries is a museum in the centre of town. Their post code is cb23bj.
    
    \textbf{User:} Great! I am also looking for a place to eat in the same area. Something not too expensive, but not cheap.
    \end{minipage}}
    \caption{Dialogue \textit{PMUL4258} from the MultiWOZ $2.1$ corpus.}
    \label{fig:dial_eg0}
\end{figure}
\begin{figure}[h!]
    \centering
    \fbox{\begin{minipage}{19em}
    \textbf{User:} Hi, I am looking for a hotel by the name of Acorn guest house.
    
    \textbf{System:} Sure, what would you like to know about it?
    
    \textbf{User:} I would like to know if it is available for 8 people for 4 nights starting Saturday?
    \end{minipage}}
    \caption{Dialogue \textit{PMUL4605} from the MultiWOZ $2.1$.}
    \label{fig:dial_eg}
\end{figure}
\begin{figure}[h!]
    \centering
    \fbox{\begin{minipage}{19em}
    \textbf{User:} I need a train to Stansted airport that leaves on Sunday.
    
    \textbf{System:} Did you have a time you would like to arrive or leave?
    
    \textbf{User:} I need to arrive by 14:30.
    
    \textbf{System:} tr1668 will arrive at 14:08, would that work for you?
    
    \textbf{User:} That is perfect. I would like to make a booking for 6 people please.
    
    \textbf{System:} Booking was successful, the total fee is 48.48 gbp payable at the station. Your reference number is hf03ug02. Do you need assistance with anything else?
    
    \textbf{User:} I need to eat too.
    
    \textbf{System:} What type of restaurant and price range are you looking for?
    
    \textbf{User:} I would like Catalan food. It needs to be in the centre and be expensive.
    \end{minipage}}
    \caption{Dialogue \textit{PMUL3625} from the MultiWOZ $2.1$ corpus.}
    \label{fig:dial_eg2}
\end{figure}

\end{document}